\documentclass{article}

    \PassOptionsToPackage{numbers, compress}{natbib}

    \usepackage[preprint]{neurips_2025}

\usepackage[utf8]{inputenc} 
\usepackage[T1]{fontenc}    
\usepackage{hyperref}       
\usepackage{url}            
\usepackage{booktabs}       
\usepackage{amsfonts}       
\usepackage{nicefrac}       
\usepackage{microtype}      
\usepackage{graphicx}
\usepackage{bm}
\usepackage{amsmath}
\usepackage{threeparttable}
\usepackage{multirow}
\usepackage[table,xcdraw]{xcolor}
\usepackage{longtable}
\usepackage{booktabs} 
\usepackage{wrapfig}
\usepackage{listings}
\usepackage[normalem]{ulem}

\title{FreeA: Plug-and-play Human-object Interaction Detection with Free Labeling}

\author{%
  Qi Liu, Yuxiao Wang\thanks{Corresponding author.}, Xinyu Jiang, Wolin Liang, Zhenao Wei, Yu Lei, Nan Zhuang, Weiying Xue\\
  School of Future Technology\\
  South China University of Technology\\
  Guangdong, GuangZhou 511400 \\
}

\begin{document}

\maketitle

\begin{abstract}
  Recent human-object interaction (HOI) detection methods depend on extensively annotated image datasets, which require a significant amount of manpower. In this paper, we propose a novel self-adaptive, language-driven HOI detection method, termed FreeA. This method leverages the adaptability of the text-image model to generate latent HOI labels without requiring manual annotation. Specifically, FreeA aligns image features of human-object pairs with HOI text templates and employs a knowledge-based masking technique to decrease improbable interactions. Furthermore, FreeA implements a proposed method for matching interaction correlations to increase the probability of actions associated with a particular action, thereby improving the generated HOI labels. Experiments on two benchmark datasets showcase that FreeA achieves state-of-the-art performance among weakly supervised HOI competitors. Our proposal gets +\textbf{13.29} (\textbf{159\%$\uparrow$}) mAP and +\textbf{17.30} (\textbf{98\%$\uparrow$}) mAP than the newest ``Weakly'' supervised model, and +\textbf{7.19} (\textbf{28\%$\uparrow$}) mAP and +\textbf{14.69} (\textbf{34\%$\uparrow$}) mAP than the latest ``Weakly+'' supervised model, respectively, on HICO-DET and V-COCO datasets, more accurate in localizing and classifying the interactive actions. The source code will be made public.
\end{abstract}

\section{Introduction}
\label{Introduction}

Human-object interaction (HOI) aims to localize and classify the interactive actions between a human and an object, enabling a more advanced understanding of images~\cite{chao2018learning}. Specifically, the HOI detection task involves taking an image as input to generate a series of triplets ($\langle$``human'', ``interaction'', ``object'' $\rangle$). Consequently, the success of this task is mainly attributed to the accurate localization of human and object entities, correct classification of object categories, and precise delineation of interaction relationships between humans and objects.

\begin{figure}[!ht]
\centering 
\includegraphics[width=\linewidth]{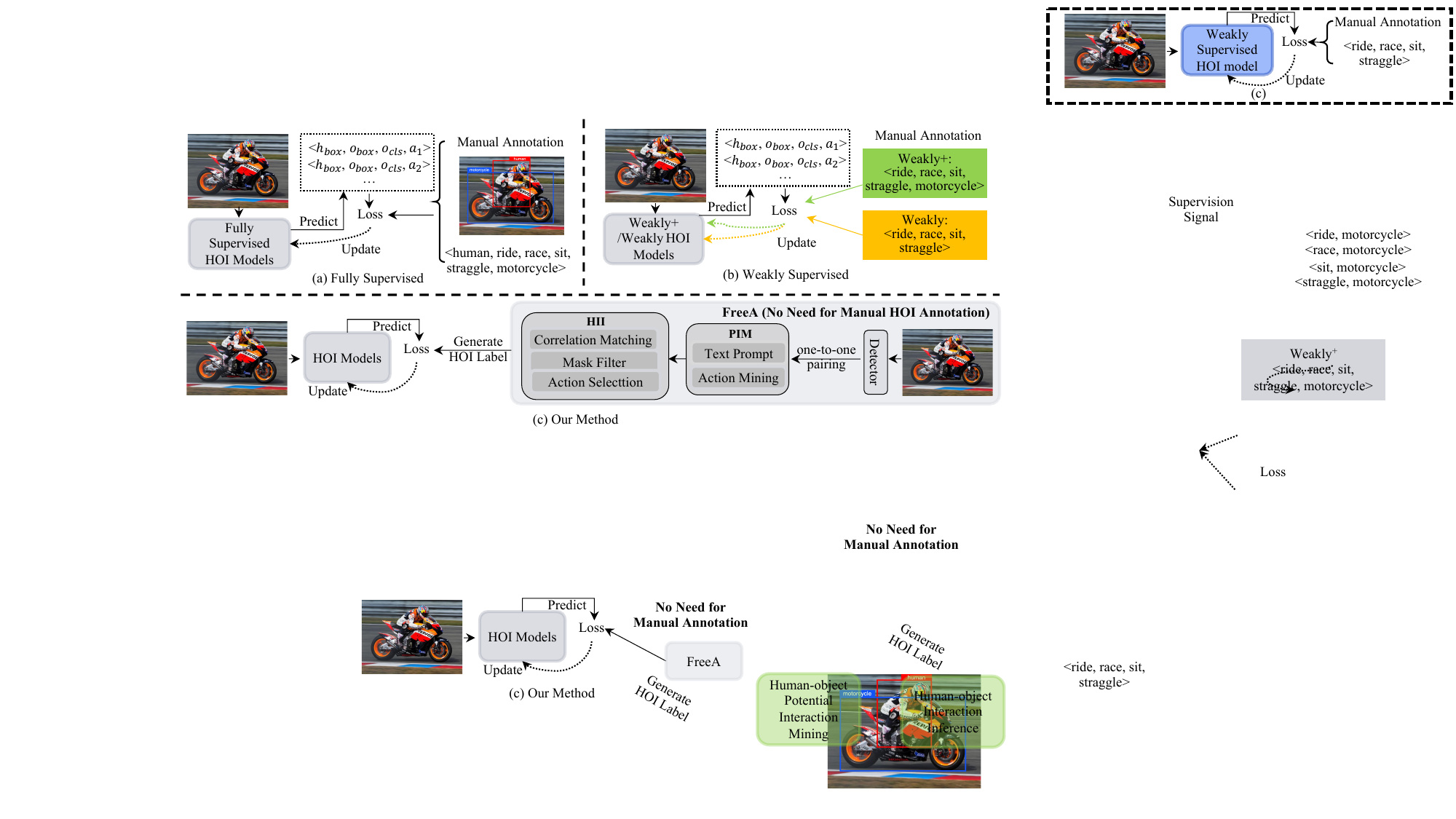}
\caption{HOI methods overview: (a) Fully supervised HOI models. Labels consist of human bounding boxes, object bounding boxes, object categories, and interaction actions of each human-object pair. (b) HOI weakly supervised. It divides into twofolds: weakly+ (using $\langle$``interaction'', ``object''$\rangle$ labels),  and weakly (using $\langle$``interaction''$\rangle$ labels). (c) Our method, i.e., FreeA, automatically generates HOI labels for HOI model training without the need for any manual annotation.}
\label{fig:figure_1}
\vspace{-0.4cm}
\end{figure}

Whether one-stage~\cite{zou2021end,liao2022gen,yuan2023rlipv2,wang2024ted} or two-stage~\cite{iftekhar2021gtnet,liu2020consnet,tip-9547056,tip-9489275} HOI detection models, they predominantly rely on computationally heavy training and requires extensive-annotation datasets (Figure \ref{fig:figure_1}(a) and Figure \ref{fig:figure_1}(b)). Taking the train set of the HICO-Det dataset as an example, even for weakly or weakly+ supervised approaches, it still needs to annotate 117,871 interaction labels from (117,871$\times$600) or (117,871$\times$23) potential combinations. It is quite resource-intensive. 
Current weakly supervised HOI models can be divided into two folds (as shown in Figure \ref{fig:figure_1}(b)): 
weakly+ using $\langle$``interaction'', ``object''$\rangle$ label ~\cite{kumaraswamy2021detecting,wan2023weakly}, e.g., eat-banana, and weakly with only $\langle$``interaction''$\rangle$ labels, e.g., eat.
Both still require abundant annotations for large-scale datasets to achieve satisfactory performance.

Another sub-direction of HOI detection is zero-shot HOI detection \cite{hou2021affordance,hou2021detecting,liao2022gen,ning2023hoiclip}, where models train on a subset of annotated data but test to unseen interaction categories. Although it aims to identify novel samples, it still depends on manually annotated training data.

Inspired by the effectiveness of the text-image matching model, e.g., CLIP~\cite{radford2021learning}, BLIP~\cite{li2022blip}, and BLIP2~\cite{li2023blip}, in accurately pairing with a given image, FreeA is designed to accomplish HOI detection without relying on manual labels (Figure \ref{fig:figure_1}(c)). 
The FreeA mainly comprises three-folds: candidate image construction (CIC), human-object potential interaction mining (PIM), and human-object interaction inference (HII). In the phase of CIC, FreeA is plug-and-play, applying existing object detection methods for all potential instances localization, and utilizes spatial denoising and pairing techniques to establish candidate interaction pairs within the image. The PIM module extensively leverages the adaptability of the text-image model in the target domain to align the high-dimensional image features with HOI interaction templates, which generates the similarity vectors of candidate interaction relationships. Then, the HII module combines the resulting similarity vectors with a priori HOI action masks to mitigate interference from irrelevant relationships, and augments the likelihood of specific actions via the proposed interaction correlation matching method. Moreover, an adaptive threshold in HII generates HOI labels for training dynamically.


Our key contributions are summarized as three-folds:

\begin{itemize}
\item [1)]
We propose a novel HOI detection method, namely FreeA, that automatically generates HOI labels. To the best of our knowledge, it is the first framework to successfully achieve HOI detection tasks without the need for manual labeling.

\item [2)] HOI detection includes various interactions among multiple instances. Three key challenges are required to tackle when generating labels, namely, multiple actions selection, filtering out irrelevant actions, and refining the coarse text-image matching's coarse labels. To address them, three corresponding modules are presented that significantly improve the effectiveness of the localization and classification of the interaction.

\item [3)]
A broad variety of experiments are conducted to demonstrate the remarkable results of the proposal on the HOI detection task, and ours performs the best among all weakly+, weakly, and fully supervised HOI models.
\end{itemize}

\section{Related Work}
\label{Related Work}

\textbf{Supervised HOI Detection.} Supervised HOI models train their networks with the help of manually annotation $\langle$``human'', ``object'', ``action''$\rangle$. The two-stage methods first use pre-trained object detection network~\cite{ren2015faster,girshick2015fast} to detect humans and objects, and then pair them one by one into the interactive discrimination network to achieve HOI detection~\cite{iftekhar2021gtnet,liu2020consnet,tip-9547056,tip-9489275}. 
However, it is pretty inefficient for one-to-one pairing between humans and objects~\cite{liao2020ppdm}. To address that, one-stage HOI detection based on transformer is gradually developed via end-to-end solution ~\cite{zou2021end,liao2022gen,wang2024ted,yuan2023rlipv2}. In addition, it has been verified that text information enables to improve the HOI detection performance~\cite{li2022improving,liao2022gen}. Current HOI approaches, e.g., GEN-VLKT~\cite{liao2022gen} and TED-Net~\cite{wang2024ted}, took use of CLIP~\cite{radford2021learning} to train the encoding network via image-text pairs.

\begin{figure*}[ht]
\centering 
\includegraphics[width=\linewidth]{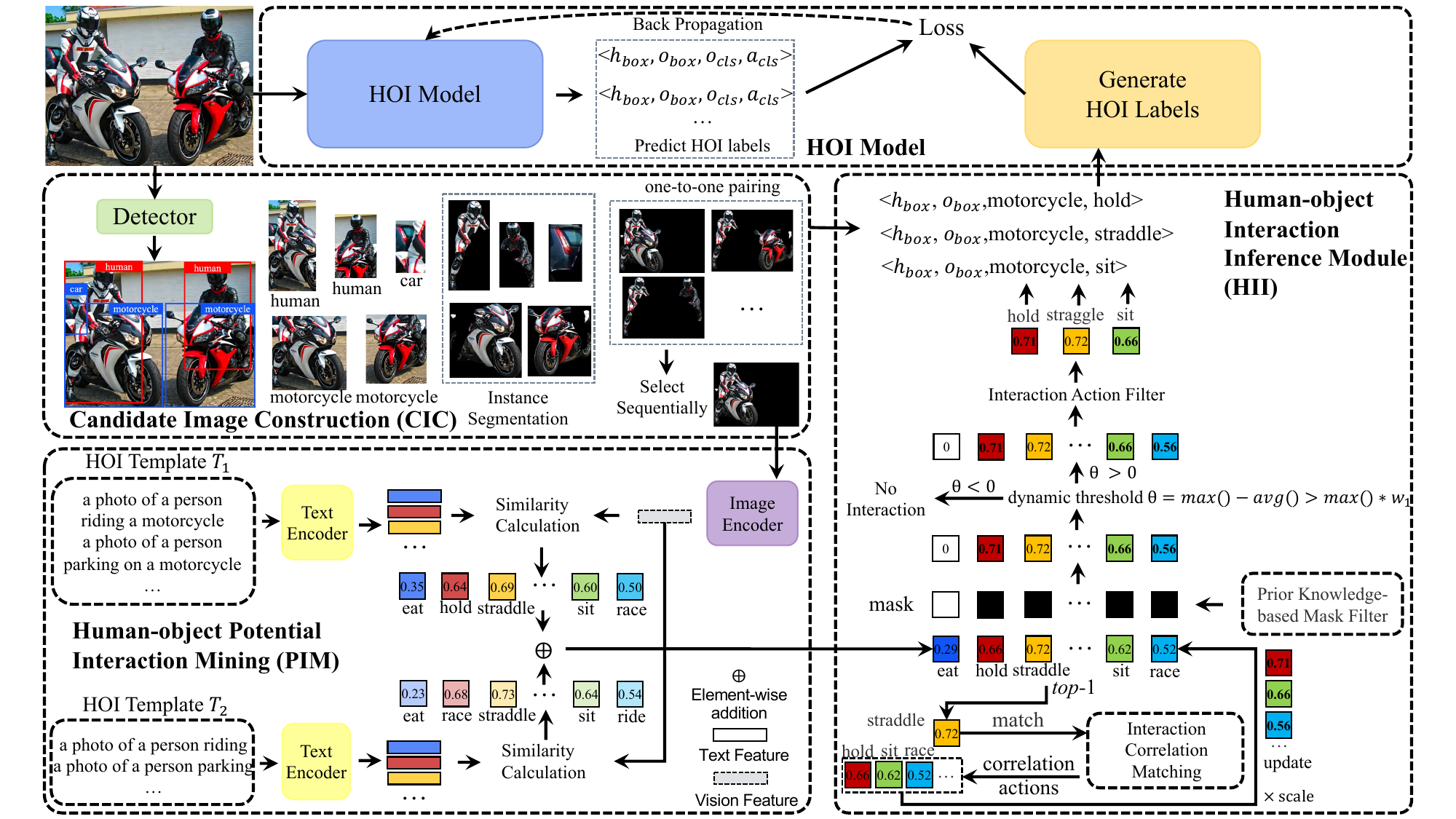}
\caption{Method overview. Starting from an existing HOI model, we apply the candidate image construction to extract humans and objects by detection and segmentation, and establish one-to-one human-object pairing. The human-object potential interaction mining module gets initial HOI interaction labels from candidate image pairs, and uses text-image matching model for domain adaptation. The human-object interaction inference module further refines these interaction labels by using $a$ $prior$ knowledge-based mask to eliminate implausible actions and using the interaction correlation matching method to enhance relevant action similarity. Finally, HOI labels are generated for model training through a dynamic threshold selector and interaction action filter.
}
\label{fig:Network}
\vspace{-0.4cm}
\end{figure*}

\textbf{Weakly Supervised HOI Detection.} Weakly supervised HOI detection generally uses image-level interaction labels for training~\cite{kilickaya2021human,kumaraswamy2021detecting,unal2023weakly}.
They can be divided into twofolds: ``weakly+'' with $\langle$``interaction'', ``object''$\rangle$ annotations~\cite{kilickaya2021human,kumaraswamy2021detecting,wan2023weakly}, and ``weakly'' with only $\langle$``interaction''$\rangle$ annotations~\cite{unal2023weakly}. MX-HOI~\cite{kumaraswamy2021detecting} proposed a momentum-independent learning framework using weakly+ supervised ($\langle$``interaction'', ``object''$\rangle$). Align-Former~\cite{kilickaya2021human} proposed an ``align layer'' to achieve pseudo alignment for training based on transformer architecture. Nevertheless, these methods suffer from noisy human-object association and ambiguous interaction types~\cite{wan2023weakly}. Therefore, Weakly HOI-CLIP~\cite{wan2023weakly} used CLIP as an interactive prompt network and HOI knowledge to enhance interaction judgment at the instance level. VLHOI~\cite{unal2023weakly} applied a language model to query for unnecessary interaction pairing reduction with image-level $\langle$``interaction''$\rangle$ annotations. Both of them, however, are still dependent on pre-annotated datasets at the cost of manpower.
Unlike the latest methods~\cite{unal2023weakly} and~\cite{wan2023weakly}, we propose a method for HOI detection that does not require manual annotation.

\textbf{Zero-shot HOI Detection.} The goal of zero-shot HOI detection is to train on a subset of labels and test using another set of unseen labels to detect interactions that were not encountered during training. Many methods~\cite{hou2021affordance,hou2021detecting,shen2018scaling,ulutan2020vsgnet,liao2022gen,ning2023hoiclip,eum2021semantics} are investigated to handle zero-shot HOI detection. Shen et al.~\cite{shen2018scaling} pioneered the application of zero-shot learning methods to address the long-tail problem in HOI detection. 
GEN-VLKT~\cite{liao2022gen} is a simple yet effective framework that utilizes CLIP for knowledge transfer~\cite{ning2023hoiclip,unal2023weakly}, thereby discovering unknown samples. The introduction of zero-shot learning enhances the adaptability of these methods to real-world scenarios. However, these methods still require manually annotated complete HOI labels. 

\vspace{-0.3cm}
\section{Method}
As shown in Figure \ref{fig:Network}, we propose a novel plug-and-play HOI framework, namely, FreeA, to reduce the requirements of annotations. The proposed framework includes candidate image construction, human-object potential relationship mining, and human-object relationship inference. Details are introduced below.

\subsection{Candidate Image Construction}
\label{sec:3_2}

To achieve automatic labels generation, we need to accurately localize humans and objects before, where Yolov8\footnote{https://github.com/ultralytics/ultralytics} model is used for localization. Given an input image $\bm{I}$, we obtain a collection of instance bounding boxes $\bm{B} = \{(\bm{b}_i | i = 1, 2, ..., N)\}$, where $\bm{b}_i = (x_i, y_i, W_i, H_i, c_i)$ from Yolov8, with $x_i$ and $y_i$ representing the center coordinates of the $i$th bounding box. $W_i$ and $H_i$ are the width and height of the bounding box, and $c_i$ denotes the category of the instance within the box. Here, $N$ represents the number of detected instances.

Subsequently, we trim the images using the bounding boxes and then pair humans with objects to create candidate images. $N_I$ candidate images are obtained, where $N_I = N_h \times N_o$, and $N_h + N_o = N$. $N_h$ and $N_o$ represent the number of humans and objects, respectively. To eliminate interference from redundant instances and background information, we apply instance segmentation to assist the PIM module in focusing on the interest of interactions.

\subsection{Human-object Potential Interaction Mining}
\label{sec:3_3}
It aims to mine potential relationships between humans and objects. We used the CLIP model to initialize the text encoder and image encoder. It is used to transfer knowledge from the source domain to the target domain. Then by computing cross-modal similarity, text-image pairs are matched.


\textbf{Image Encoder}. The image encoder $\mathbf{F}_{IE}$ is employed to process the set of $N_I$ candidate images, with the result of image encoding $\bm{I}_E \in \mathbb{R}^{N_I \times C_{IE}}$. The $\bm{C}_{IE}$ denotes the dimension of the image encoder, and $\bm{I}_B$ represents the collection of candidate images. We have:
\begin{equation}
\label{eq:1}
        \bm{I}_E=\mathbf{F}_{IE}(\bm{I}_B).
\end{equation}
\ \ \ \textbf{Text Encoder}. We start with a text template creation, denoted as $\bm{T}_1$, in the format ``a photo of a person verb-ing an object''. For example, the triplet $\langle$ ``human'', ``ride'', ``motorcycle''$\rangle$ is transformed into ``a photo of a person riding a motorcycle''. After that, $\bm{T}_1$ is input into the text encoder $\mathbf{F}_{TE}$, leading to a text information matrix $\bm{T}_{E1}$ $\in \mathbb{R}^{N_T \times C_{TE}}$. \(N_T\) denotes the number of textual queries, i.e., the number of HOI interaction categories. 
For the HICO-Det dataset, \(N_T = 600\); for V-COCO, \(N_T = 284\). 
\(C_{TE}\) represents the dimensionality of the encoded textual features.



To emphasize the importance of verbs in HOI relationships, another type of text template $\bm{T}_2$ in the format ``a photo of a person verb-ing'', has been constructed. $\bm{T}_1$ and $\bm{T}_2$ are distinct text templates for different HOI actions, corresponding to information matrices $\bm{T}_{E1}$ and $\bm{T}_{E2}$, respectively. They are formulated as:
\begin{equation}
    \bm{T}_{Ei}=\mathbf{F}_{TE}\left(\bm{T}_i\right),i=1,2.
\end{equation}
Next, we calculate the cosine similarity between the image encoding information $\bm{I}_E$ and the text information $\bm{T}_E$. That is:
\begin{gather}
\label{eq:cos}
    {\bm{sim}}_{Ei}(\bm{I}_E, \bm{T}_{Ei}) = \frac{\bm{I}_E \cdot \bm{T}_{Ei}^T}{\|\bm{I}_E\| \cdot \|\bm{T}_{Ei}\|}, i=1,2, \\
    \bm{S}={\bm{sim}}_{E1}+{\bm{sim}}_{E2},
\end{gather}
where $\bm{I}_E \in \mathbb{R}^{N_I \times C_{IE}}$, $\bm{T}_{Ei} \in \mathbb{R}^{N_T \times C_{TE}}$, $C_{IE}=C_{TE}$, $\bm{sim}_{Ei} \in \mathbb{R}^{N_I \times N_T}$, and $\bm{S} \in \mathbb{R}^{N_I \times N_T}$. 
\subsection{Human-object Interaction Inference}
\label{sec:3_4}

The HII module consists of interaction correlation matching, prior knowledge-based mask filter, dynamic threshold selector, and interaction action filter.

\begin{wrapfigure}{r}{0.5\textwidth}
  \centering 
\includegraphics[width=\linewidth]{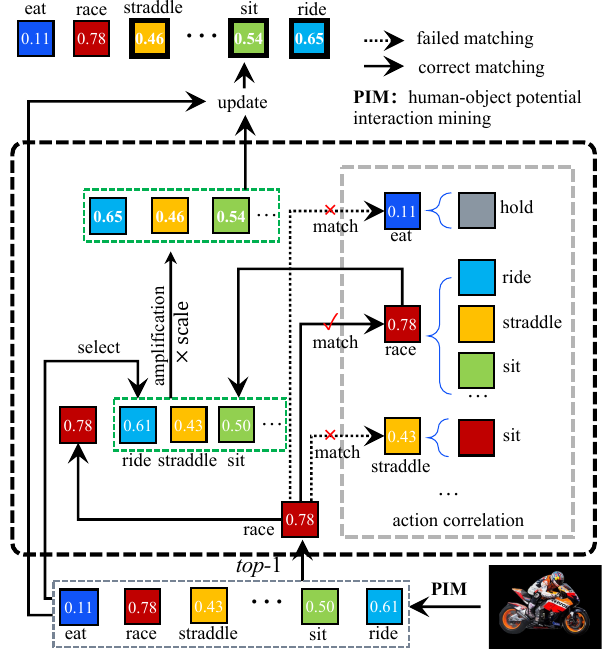}
\caption{Details of the interaction correlation matching method. 
}
\label{fig:ICM}
\vspace{0.5cm}
\end{wrapfigure}

\textbf{Interaction Correlation Matching (ICM)}. If a certain action occurs, other actions may also occur concurrently. For instance, when a person ``racing a motorcycle", he is also ``riding and sitting on the motorcycle". Inspired by that, we propose interaction correlation matching to infer other behaviors strongly correlated with the $top\text{-}1$ selected initial interaction action, as shown in Figure \ref{fig:ICM}. When ``race" is selected based on the highest similarity, we will also extract highly correlated actions, such as ``ride", ``straddle", ``sit", etc. The actions that are highly relevant to the target action 
are precomputed, and selecting the related actions is similar to a dictionary operation. Afterward, we amplify the similarity of the selected action and update the similarity vector. To be specific, for each row vector in $\bm{S}$, we employ a $top\text{-}1$ selection strategy to choose the initial interaction action with the highest image-text similarity. Furthermore, we amplify image-text similarity to highlight the similarity between these interaction actions. Detailed steps for applying interaction correlation matching to a single candidate image are:
\begin{equation}
    {\bm{k}}_a=\left[j_1,j_2,...,j_n\right]=\mathbf{F}_{ICM}(\bm{k}_{max}(\bm{S}_i)),
\end{equation}
\begin{equation}
\label{eq:scale}
    \bm{S}_{ij}=\bm{S}_{ij} \times scale,\ j\in{\bm{k}}_a,
\end{equation}
where $\bm{k}_{max}(\bm{S}_i)$ represents the index of interaction action with the highest image-text similarity in the $i$th candidate image, the function \(\mathbf{F}_{ICM}\) is a preprocessed dictionary that outputs a set of actions highly related to a given action category. $\bm{k}_a$ denotes a set of indexes associated with several interaction actions correlated with $\bm{k}_{max}(\bm{S}_i)$, and $scale$ denotes the amplification factor.

\textbf{Prior Knowledge-based Mask (PKM) Filter}. As widely acknowledged, specific objects often exhibit clear associations with particular action categories. For instance, common interaction actions with an ``apple'' include ``pick'' and ``eat'', while ``ride'' or ``drive'' are highly unlikely. Inspired by that we design \textit{a priori} knowledge-based mask filter, which uses a specific mask mechanism to filter interaction actions for specific objects based on prior knowledge. The similarity score $\bm{S}_{ij}$ is updated as:
\begin{equation}
{\bm{k}}_o=\left[j_1,j_2,...,j_n\right]=\mathbf{F}_{PKM}(\bm{o}_i),
\end{equation}
\begin{equation}
    \bm{mask} =[m_1,m_2,\ldots,m_j,\ldots,m_{N_T}]  =\left\{\begin{matrix}
 0,j\notin \bm{k}_o \\ 
 1,j\in \bm{k}_o
\end{matrix}\right.,j=1,2,…,N_T,
\end{equation}
\begin{equation}
\bm{S}_{ij}=\bm{S}_{ij} \times \bm{mask}_j,j=1,2,...,N_T,
\end{equation}
where $\bm{o}_i$ represents the object category in the $i$th image. The function \(\mathbf{F}_{\text{PKM}}\) is a preprocessed dictionary that outputs all possible actions, denoted as $\bm{k}_o$, associated with a given object category. $m_j$ = \{1, 0\}: 1 indicates that the $j$th action is related to $\bm{o}_i$, not the other way around. Therefore, we retain interaction actions related to specific object categories after reducing interference from unlikely actions. 
\begin{table*}[h]
\centering
\caption{Performance comparisons on HICO-Det dataset. * represents the results given in ~\cite{wan2023weakly,unal2023weakly}. Swin-L represents the Swin-Transformer Large model. }
\label{tab-hico-det}
\setlength{\tabcolsep}{1.3mm}{
\begin{tabular}{llllccc}
\specialrule{1.5pt}{0pt}{0pt}
\multirow{2}{*}{Methods}                                          &   \multirow{2}{*}{Object Detector}    & \multirow{2}{*}{Backbone} & \multicolumn{1}{l|}{\multirow{2}{*}{Source}} & \multicolumn{3}{c}{Default (\textbf{mAP}$\uparrow$)}                                                         \\
                                                   &                       &                           & \multicolumn{1}{l|}{}                          & \multicolumn{1}{l}{Full} & \multicolumn{1}{l}{Rare} & \multicolumn{1}{l}{None-Rare} \\ \hline
\multicolumn{6}{c}{fully supervised (using \textcolor{red}{$\langle$``human", ``interaction", ``object" $\rangle$} labels)}                                                                                                                                                                                                                                                                                                \\
InteractNet~\cite{gkioxari2018detecting}   & Faster R-CNN        & ResNet-50-FPN             & \multicolumn{1}{l|}{CVPR}                 & 9.94                     & 7.16                     & \multicolumn{1}{c}{10.77}     \\
iCAN~\cite{gao2018ican}           & Faster R-CNN                & ResNet-50                 & \multicolumn{1}{l|}{BMCV}                 & 14.84                    & 10.45                    & \multicolumn{1}{c}{16.15}     \\
PMFNet~\cite{wan2019pose}         & Faster R-CNN                 & ResNet-50-FPN             & \multicolumn{1}{l|}{ICCV}                 & 17.46                    & 15.56                    & \multicolumn{1}{c}{18.00}     \\
DJ-RN~\cite{li2020detailed}        & Faster R-CNN                & ResNet-50                 & \multicolumn{1}{l|}{CVPR}                 & 21.34                    & 18.53                    & \multicolumn{1}{c}{22.18}     \\
IDN~\cite{li2020hoi}              & Faster R-CNN                & ResNet-50                 & \multicolumn{1}{l|}{NeurIPS}              & 23.36                    & 22.47                    & \multicolumn{1}{c}{23.63}     \\
QPIC~\cite{tamura2021qpic}        & DETR                 & ResNet-101                & \multicolumn{1}{l|}{CVPR}                 & 29.90                    & 23.92                    & \multicolumn{1}{c}{31.69}     \\
RCL~\cite{kim2023relational}       & DETR                & ResNet-50                 & \multicolumn{1}{l|}{CVPR}                 & 32.87                    & 28.67                    & \multicolumn{1}{c}{34.12}     \\

TED-Net~\cite{wang2024ted}        & DETR             & ResNet-50                 & \multicolumn{1}{l|}{TCSVT}                 & 34.00                    & 29.88                    & \multicolumn{1}{c}{35.24}        \\ 
PViC~\cite{zhang2023exploring}     & DETR                & ResNet-50                 & \multicolumn{1}{l|}{ICCV}                 & 34.69                    & 32.14                    & \multicolumn{1}{c}{35.45}        \\ 
PViC~\cite{zhang2023exploring}       & DETR                & Swin-L                 & \multicolumn{1}{l|}{ICCV}                 & \textbf{44.32}                    & 44.61                    & \multicolumn{1}{c}{44.24}        \\
RLIPv2~\cite{yuan2023rlipv2}        & DETR             & Swin-L                 & \multicolumn{1}{l|}{ICCV}                 & 43.23                    & \textbf{45.64}                    & \multicolumn{1}{c}{\textbf{45.09}}        \\ \hline
\multicolumn{6}{c}{weakly+ supervised (using \textcolor{red}{$\langle$``interaction", ``object" $\rangle$} labels)}                                                                                                                                                                                                                                                                                             \\
Explanation-HOI*~\cite{baldassarre2020explanation}   & Faster R-CNN & ResNeXt-101                & \multicolumn{1}{l|}{ECCV}                 & 10.63                    & 8.71                     & \multicolumn{1}{c}{11.20}    \\
MAX-HOI~\cite{kumaraswamy2021detecting}      & Faster R-CNN        & ResNet-101                 & \multicolumn{1}{l|}{WACV}                 & 16.14                    & 12.06                    & \multicolumn{1}{c}{17.50}    \\
Align-Former~\cite{kilickaya2021human}        & DETR        & ResNet-101                & \multicolumn{1}{l|}{arXiv}                & 20.85                    & 18.23                    & \multicolumn{1}{c}{21.64}   \\
PPR-FCN*~\cite{zhang2017ppr}               & R-FCN         & ResNet-50                 & \multicolumn{1}{l|}{ICCV}                 & 17.55                    & 15.69                    & \multicolumn{1}{c}{18.41}    \\
Weakly HOI-CLIP~\cite{wan2023weakly}                & Faster R-CNN          & ResNet-50                 & \multicolumn{1}{l|}{ICLR}                 & 22.89                    & 22.41                    & \multicolumn{1}{c}{23.03}      \\
\rowcolor[gray]{0.9} Ours                    & Faster R-CNN                                                 & ResNet-50                 & \multicolumn{1}{c|}{-}                          & 24.14                    & 20.91                    & \multicolumn{1}{c}{24.09}        \\
OpenCat~\cite{zheng2023open} & Faster R-CNN & ResNet-101 & \multicolumn{1}{l|}{CPVR} & \multicolumn{1}{c}{25.82} & 24.35 & 26.19 \\
\rowcolor[gray]{0.9} Ours                    & Faster R-CNN                                                 & ResNet-101                 & \multicolumn{1}{c|}{-}                          & 27.02                    & 24.49                    & \multicolumn{1}{c}{28.37}        \\
\rowcolor[gray]{0.9} Ours                    & DETR                                                 & ResNet-50                 & \multicolumn{1}{c|}{-}                          & 24.33                    & 21.13                    & \multicolumn{1}{c}{24.77}        \\

\rowcolor[gray]{0.9} Ours                    & Yolov8                                                 & ResNet-50                 & \multicolumn{1}{c|}{-}                          & 24.57                    & 21.45                    & \multicolumn{1}{c}{25.51}        \\
\rowcolor[gray]{0.9} Ours                & Yolov8                                                     & Swin-L                 & \multicolumn{1}{c|}{-}                          & \textbf{\textbf{33.01}}                    & \textbf{32.13}                    & \multicolumn{1}{c}{\textbf{33.27}}        \\ \hline
\multicolumn{6}{c}{weakly supervised (using \textcolor{red}{$\langle$``interaction" $\rangle$} labels)}                                                                                                                                                                                                                                                                                              \\
SCG*~\cite{zhang2021spatially}              & Faster R-CNN      & ResNet-50                 & \multicolumn{1}{l|}{ICCV}                 & 7.05                     & -                        & \multicolumn{1}{c}{-}       \\
VLHOI~\cite{unal2023weakly}                            & Faster R-CNN                                     & ResNet-50                 & \multicolumn{1}{l|}{CVPR}                & 8.38                     & -                        & \multicolumn{1}{c}{-}        \\ 
\rowcolor[gray]{0.9}Ours (no use labels)                & Faster R-CNN                                                     & ResNet-50                 & \multicolumn{1}{c|}{-}                         & 15.85                    & 15.11                          & \multicolumn{1}{c}{16.69}        \\
\rowcolor[gray]{0.9}Ours (no use labels)                & Yolov8                                                     & ResNet-50                 & \multicolumn{1}{c|}{-}                         & 16.96                    & 16.26                          & \multicolumn{1}{c}{17.17}        \\
\rowcolor[gray]{0.9}Ours (no use labels)        & Yolov8                                                             & Swin-L                 & \multicolumn{1}{c|}{-}                         & \textbf{21.67}                    & \textbf{23.69}                          & \multicolumn{1}{c}{\textbf{21.06}}        \\ \specialrule{1.5pt}{0pt}{0pt}
\end{tabular}
}
\end{table*}

\textbf{Dynamic Threshold Selector}. A dynamic threshold selector is employed to assess whether interaction has occurred in a candidate image. Our starting point is that when the difference between the maximum value and the mean value of $\bm{S}_i$ is greater than the maximum value of $\bm{S}_i$ multiplied by a weighting factor, it is considered that the maximum value of $\bm{S}_i$ has a significant gap with the other values. This indicates that there may be an interaction (when there is interaction in the image, the similarity of the specific action tends to be high, while the similarity of unrelated actions tends to be low). The calculation formula is written as:
\begin{equation}
\label{eq:dynamic-weight}
    \theta=(max\left(\bm{S}_i\right)-\frac{\sum_{j=1}^{N_T}\bm{S}_{ij}}{N_T})-max(\bm{S}_i)\times\omega_1,
\end{equation}
where $max(\bm{S}_i)$ is the highest image-text similarity value in the $i$th candidate image, and $\omega_1$ is a parameter to balance the threshold range. A positive $\theta$ ($\theta>0$) indicates the presence of human-object interaction in $i$th candidate image. This dynamic threshold adjustment enhances the accuracy of interaction relationship detection and recognition for different scenarios, further leading to more precise event determination.

\textbf{Interaction Action Filter}. In the presence ($\theta > 0$) of interaction in the candidate image, we employ an interaction action filtering to select target actions from $N_T$ interaction actions. The filtering procedure is expressed as:
\begin{equation}
\label{eq:adapt}
\bm{a}_{index}=\{j|(max\left(\bm{S}_i\right)-max\left(\bm{S}_i\right)\times\omega_2)<\bm{S}_{ij}<max\left(\bm{S}_i\right),\bm{S}_{ij}\in \bm{S}_i\},
\end{equation}
where $j$ represents the index of each interaction action, and $\bm{a}_{index}$ denotes the set of action indices. Finally, the HOI labels $\mathcal{O}$ are built as:
\begin{equation}
\mathcal{O}=\{(\bm{h}_{box},\bm{o}_{box},c_o,\bm{a}_i)|\bm{a}_i\in \bm{a}_{index}\},
\end{equation}
where $\bm{h}_{box}$ and $\bm{o}_{box}$ are the detected bounding boxes of the human and object entities, respectively. $c_o$ denotes the object category, and $\bm{a}_i$ represents the interaction action index.

The total loss function is consistent with that of GEN-VLKT~\cite{liao2022gen}, defined as:
\begin{equation}
    \mathcal{L} =  \lambda_b \sum_{i\in(h,o)} \mathcal{L}_{b}^i+\lambda_u \sum_{j\in(h,o)} \mathcal{L}_{u}^j + \sum_{k\in(o,a)} \lambda_c^k \mathcal{L}_{c}^k,
\end{equation}
where $\mathcal{L}_{b}$, $\mathcal{L}_{u}$, and $\mathcal{L}_{c}$ are box regression loss, IoU loss, and classification loss, respectively. $\lambda_b$, $\lambda_u$ and $\lambda_c^k$ are the hyper-parameters for adjusting the weights of each loss.

\section{Experiments}
\label{experiments}
\textbf{Datasets}. The benchmark datasets, HICO-Det and V-COCO, are used to demonstrate the effectiveness of the proposed method. HICO-Det includes 47,776 images and covers 80 object categories, 117 action categories, and 600 distinct interaction types. V-COCO comprises 10,346 images, featuring 80 object categories and 29 action categories, including 4 body actions without object interactions. 

\begin{table}[t]
\small
  \centering
  \begin{minipage}[t]{0.48\linewidth}
    \centering
    \caption{Performance comparisons on V-COCO dataset. * represents the results given in ~\cite{unal2023weakly}. VLHOI† is trained via video-captured labels. The object detector of our is Yolov8.}
\label{tab-v-coco}
    \setlength{\tabcolsep}{0.8mm}{
    \begin{tabular}{llcc}
\specialrule{1.5pt}{0pt}{0pt}
\multirow{2}{*}{Method}                                                   & \multirow{2}{*}{Backbone} & \multirow{2}{*}{\begin{tabular}[c]{@{}l@{}}AP$^{S1}_{role}$\\ (\textbf{mAP}$\uparrow$)\end{tabular}} & \multirow{2}{*}{\begin{tabular}[c]{@{}l@{}}AP$^{S2}_{role}$\\ (\textbf{mAP}$\uparrow$)\end{tabular}} \\
                                                                          &                       &                                                                      &                                                                      \\ \hline
\multicolumn{4}{c}{weakly+ supervised}                                                                                                                                                                                                          \\
Align-Former               & ResNet-101            & \multicolumn{1}{c}{15.82}                                                                & \multicolumn{1}{c}{16.34}                                                                \\
Weakly HOI-CLIP                           & ResNet-50             & \multicolumn{1}{c}{42.97}                                                                & \multicolumn{1}{c}{48.06}                                                                \\
OpenCat & ResNet-101 & 34.40 & 36.10 \\
\rowcolor[gray]{0.9}Ours                                                                      &   \multicolumn{1}{c}{ResNet-50}                    & \multicolumn{1}{c}{50.25}                                                                & \multicolumn{1}{c}{52.05}                                                                \\
\rowcolor[gray]{0.9}Ours                                                                      &   \multicolumn{1}{c}{Swin-L}                    & \multicolumn{1}{c}{\textbf{57.66}}                                                                & \multicolumn{1}{c}{\textbf{59.78}}                                                                \\\hline
\multicolumn{4}{c}{weakly supervised}                                                                                                                                                                                                           \\
SCG*                      & ResNet-50             & \multicolumn{1}{c}{20.05}                                                                 & \multicolumn{1}{c}{-}                                                \\
VLHOI                    & ResNet-50            & \multicolumn{1}{c}{29.59}                                                                & \multicolumn{1}{c}{-}                                                \\ 
VLHOI†                    & ResNet-50            & \multicolumn{1}{c}{17.71}                                                                & \multicolumn{1}{c}{-}                                                \\ 
\rowcolor[gray]{0.9}\begin{tabular}[c]{@{}l@{}}Ours (no labels)\end{tabular}                                                                      & \multicolumn{1}{c}{ResNet-50} & \multicolumn{1}{c}{30.82}                                                                & \multicolumn{1}{c}{32.60}                                                                \\
\rowcolor[gray]{0.9}\begin{tabular}[c]{@{}l@{}}Ours (no labels)\end{tabular}                                                                      & \multicolumn{1}{c}{Swin-L} & \multicolumn{1}{c}{\textbf{35.01}}                                                                & \multicolumn{1}{c}{\textbf{37.30}}                                                                \\ \specialrule{1.5pt}{0pt}{0pt}
\end{tabular}}
  \end{minipage}
  \hfill
  \begin{minipage}[t]{0.48\linewidth} 
    \centering
    \caption{Performance comparisons on V-COCO dataset with different object detectors and backbones.}
    \label{tab-v-coco-different-detector}
    \setlength{\tabcolsep}{0.4mm}{
    \begin{tabular}{llll}
\specialrule{1.5pt}{0pt}{0pt}
\multirow{2}{*}{Method}                                                   & \multirow{2}{*}{\begin{tabular}[c]{@{}c@{}}Object\\ Detector\end{tabular}} & \multirow{2}{*}{Backbone} & \multirow{2}{*}{\begin{tabular}[c]{@{}l@{}}AP$^{S1}_{role}$\\ (\textbf{mAP}$\uparrow$)\end{tabular}} \\
                                                                          &                       &                                                                      &                                                                      \\ \hline
\multicolumn{4}{c}{weakly+ supervised}                                                                                                                                                                                                          \\
Weakly HOI-CLIP                        &  Faster R-CNN   & ResNet-50             & \multicolumn{1}{c}{42.97}                                                                                                                                \\
OpenCat                       &  Faster R-CNN   & ResNet-101             & \multicolumn{1}{c}{34.40}                                                                                                                                \\
\rowcolor[gray]{0.9}Ours        &      Faster R-CNN                                                        &   \multicolumn{1}{c}{ResNet-50}                    & \multicolumn{1}{c}{46.32}                                                                                                                               \\
\rowcolor[gray]{0.9}Ours        &      DETR                                                        &   \multicolumn{1}{c}{ResNet-50}                    & \multicolumn{1}{c}{48.11}                                                                                                                               \\
\rowcolor[gray]{0.9}Ours        &      Yolov8                                                        &   \multicolumn{1}{c}{ResNet-50}                    & \multicolumn{1}{c}{50.25}                                                                                                                               \\
\rowcolor[gray]{0.9}Ours       &   Yolov8                                                            &   \multicolumn{1}{c}{Swin-L}                    & \multicolumn{1}{c}{\textbf{57.66}}                                                                                                                              \\\hline
\multicolumn{4}{c}{weakly supervised}                                                                                                                                                                                                           \\
VLHOI†           &   Faster R-CNN      & ResNet-50            & \multicolumn{1}{c}{17.71}                                                                                                              \\ 
\rowcolor[gray]{0.9}\begin{tabular}[c]{@{}l@{}}Ours (no labels)\end{tabular}                                                         &     Faster R-CNN        & \multicolumn{1}{c}{ResNet-50} & \multicolumn{1}{c}{28.12}         \\
\rowcolor[gray]{0.9}\begin{tabular}[c]{@{}l@{}}Ours (no labels)\end{tabular}                                                         &     DETR        & \multicolumn{1}{c}{ResNet-50} & \multicolumn{1}{c}{29.93}         \\
\rowcolor[gray]{0.9}\begin{tabular}[c]{@{}l@{}}Ours (no labels)\end{tabular}                                                         &     Yolov8        & \multicolumn{1}{c}{ResNet-50} & \multicolumn{1}{c}{30.82}         \\
\rowcolor[gray]{0.9}\begin{tabular}[c]{@{}l@{}}Ours (no labels)\end{tabular}                                                         &    Yolov8         & \multicolumn{1}{c}{Swin-L} & \multicolumn{1}{c}{\textbf{35.01}}                                                                                                                              \\ \specialrule{1.5pt}{0pt}{0pt}
\end{tabular}}
    
  \end{minipage}
\end{table}


\subsection{Effectiveness for Regular HOI Detection}
As shown in Table~\ref{tab-hico-det}, we compare our method with various supervision levels on HICO-DET. Under weakly supervision using only ``interaction'' labels, our ResNet-50-based approach surpasses the SOTA VLHOI method by 7.47 mAP—without requiring manual annotations. With a Swin-L backbone and a YOLOv8 detector, our method achieves 21.67 mAP in the Full setting. Extending to the weakly+ setting, we exceed the latest Weakly HOI-CLIP method by 1.25 and 1.06 mAP in the Full and Non-Rare settings, respectively, using ResNet-50 and Faster R-CNN. However, performance on Rare categories is lower due to our pseudo-labeling strategy: when passed through CLIP, rare-class actions often yield low similarity scores and are thus filtered out, impacting recognition accuracy. Additionally, with ResNet-101 and Faster R-CNN, our method outperforms OpenCat across all metrics. Using a Swin-L backbone, we surpass OpenCat by 7.19, 7.78, and 7.08 mAP in the Full, Rare, and Non-Rare settings, respectively. Notably, our method also outperforms several fully supervised HOI models (e.g., InteractNet, iCAN, QPIC, RCL). 

Experimental results on the V-COCO dataset are presented in Table \ref{tab-v-coco}. The proposed FreeA is far ahead of the VLHOI† method using utilized video-captured labels, where the result increases from 17.71 mAP to 30.82 mAP in terms of AP$^{S1}_{role}$. As well, FreeA with no labels surpasses the VLHOI model that using $\langle$``interaction''$\rangle$ labels by achieving a 1.23 mAP increase at AP$^{S1}_{role}$. Moreover, FreeA achieves a 7.28 mAP increase at AP$^{S1}_{role}$ as compared to the SOTA weakly+ supervised Weakly HOI-CLIP model. As shown in Table \ref{tab-v-coco-different-detector}, under the weakly+ setting, our method outperformed Weakly HOI-CLIP by 3.35 mAP. Furthermore, we experimented with different detectors, which further improved the results. In the weakly setting, our method without labels achieved a significant improvement of 10.41 mAP.

Table~\ref{tab-zero-shot} presents the results of our proposed method under the zero-shot learning setting. RF-UC denotes the rare first setting, while NF-UC refers to the non-rare first unseen setting. UC and UV represent the unseen composition and unseen verb settings, respectively~\cite{ning2023hoiclip,xue2025towards}. In HOI, zero-shot learning involves training the model on a subset of seen interaction categories and testing it on previously unseen categories. Our method directly generates predictions for the unseen categories and then trains the HOI model, thereby enabling evaluation under zero-shot learning conditions. As shown in the table, our method achieves promising results across various zero-shot settings, demonstrating its effectiveness and generalization capability in handling previously unseen interaction combinations. 
\subsection{Ablation Studies}
\label{ablation_study}
\setlength{\parindent}{1em}

\textbf{HOI model}. Our proposed method is plug-and-play. To evaluate the effectiveness of HOI model to the FreeA, four up-to-date fully supervised HOI approaches are tested, as shown in Table \ref{tab-hoi-models}. It is observed that a better HOI model can promote the overall effect of FreeA. 

For example, the QPIC model, with a 29.07 mAP in terms of full under fully supervised, achieves a 20.18 mAP in the weakly+ supervised. When we employ a superior HOI model, such as GEN-VLKT, it reaches 24.57 mAP in the weakly+ supervised. After replacing it with the more advanced RLIPV2 model, the results were further improved, reaching 33.01 mAP, 32.13 mAP, and 33.27 mAP, respectively. To conduct ablation experiments efficiently, the upcoming tests will use GEN-VLKT as our HOI model.

\begin{table}[t]
\small
  \centering
  \begin{minipage}[t]{0.48\linewidth}
    \centering
    \caption{Performance comparison for zero-shot HOI detection on HICO-Det.}
\label{tab-zero-shot}
    \setlength{\tabcolsep}{1.4mm}{
    \begin{tabular}{ll|ccc}
\specialrule{1.5pt}{0pt}{0pt}
\multirow{2}{*}{Method}                                                                         & \multirow{2}{*}{Source}                               & \multicolumn{3}{c}{\textbf{mAP}$\uparrow$}                                                                                                                                                 \\
                                                                                                &                                                     & Full                                                    & Unseen                                                    & Seen                                                \\ \hline
HOICLIP~\cite{ning2023hoiclip}  & UC & 32.99 & 34.85 & 25.53 \\
KI2HOI~\cite{xue2025towards} & UC & 34.56 & 35.76 & 27.43 \\
Ours     & UC & \textbf{34.91} & \textbf{35.84} & \textbf{28.14}  \\ \hline
HOICLIP~\cite{ning2023hoiclip} & UV & 31.09 & 32.19 & 24.30 \\
KI2HOI~\cite{xue2025towards} & UV & 31.85 & 32.95 & 25.20 \\
Ours     & UV & \textbf{33.39} & \textbf{35.22} & \textbf{26.47} \\ \hline
HOICLIP~\cite{ning2023hoiclip} & NF-UC & 27.75 & 28.10 & 26.39 \\
KI2HOI~\cite{xue2025towards} & NF-UC & 27.77 & 28.31 & 28.89 \\
Ours     & NF-UC & \textbf{29.15} & \textbf{29.93} & \textbf{29.34}\\ \hline
HOICLIP~\cite{ning2023hoiclip} & RF-UC  & 32.99 & 34.85 & 25.53 \\
KI2HOI~\cite{xue2025towards} & RF-UC & 34.10 & \textbf{35.79} & 26.33 \\
Ours     & RF-UC & \textbf{34.86} & 35.19 & \textbf{28.04}  \\ 
 \specialrule{1.5pt}{0pt}{0pt}
\end{tabular}}
    \vspace{-0.4cm}
  \end{minipage}
  \hfill
  \begin{minipage}[t]{0.48\linewidth} 
    \centering
    \caption{Ablation study using different HOI models in weakly+ supervised on HICO-Det datasets. $\tau$ represents performance under full supervision.}
\label{tab-hoi-models}
    \setlength{\tabcolsep}{1.0mm}{
    \begin{tabular}{ll|ccc}
\specialrule{1.5pt}{0pt}{0pt}
\multirow{2}{*}{Method}                                                                         & \multirow{2}{*}{Source}                               & \multicolumn{3}{c}{Defalut (\textbf{mAP}$\uparrow$)}                                                                                                                                                 \\
                                                                                                &                                                     & Full                                                    & Rare                                                    & Non-Rare                                                \\ \hline
\begin{tabular}[c]{@{}l@{}}QPIC\end{tabular}     & \begin{tabular}[c]{@{}l@{}}CVPR\end{tabular} & \begin{tabular}[c]{@{}l@{}}20.18\\ 29.07$^{\tau}$\end{tabular} & \begin{tabular}[c]{@{}l@{}}15.82\\ 21.85$^{\tau}$\end{tabular} & \begin{tabular}[c]{@{}l@{}}21.55\\ 31.23$^{\tau}$\end{tabular} \\ \hline
\begin{tabular}[c]{@{}l@{}}STIP\end{tabular} & \begin{tabular}[c]{@{}l@{}}CVPR\end{tabular} & \begin{tabular}[c]{@{}l@{}}21.43\\ 31.60$^{\tau}$\end{tabular} & \begin{tabular}[c]{@{}l@{}}19.03\\ 27.75$^{\tau}$\end{tabular} & \begin{tabular}[c]{@{}l@{}}22.26\\ 32.75$^{\tau}$\end{tabular} \\ \hline
\begin{tabular}[c]{@{}l@{}}RCL\end{tabular}   & \begin{tabular}[c]{@{}l@{}}CVPR\end{tabular} & \begin{tabular}[c]{@{}l@{}}23.24\\32.87$^{\tau}$\end{tabular} & \begin{tabular}[c]{@{}l@{}}19.74\\ 28.67$^{\tau}$\end{tabular} & \begin{tabular}[c]{@{}l@{}}24.38\\ 34.12$^{\tau}$\end{tabular} \\ \hline
\begin{tabular}[c]{@{}l@{}}GEN-VLKT\end{tabular}    & \begin{tabular}[c]{@{}l@{}}CVPR\end{tabular} & \begin{tabular}[c]{@{}l@{}}24.57\\ 33.75$^{\tau}$\end{tabular} & \begin{tabular}[c]{@{}l@{}}21.45\\ 29.25$^{\tau}$\end{tabular} & \begin{tabular}[c]{@{}l@{}}25.51\\ 35.10$^{\tau}$\end{tabular} \\ \hline
\begin{tabular}[c]{@{}l@{}}RLIPv2\end{tabular}    & \begin{tabular}[c]{@{}l@{}}ICCV\end{tabular} & \begin{tabular}[c]{@{}l@{}}\textbf{33.01}\\ 43.23$^{\tau}$\end{tabular} & \begin{tabular}[c]{@{}l@{}}\textbf{32.13}\\ 45.64$^{\tau}$\end{tabular} & \begin{tabular}[c]{@{}l@{}}\textbf{33.27}\\ 45.09$^{\tau}$\end{tabular} \\ \specialrule{1.5pt}{0pt}{0pt}
\end{tabular}}
    
  \end{minipage}
\end{table}

\textbf{HOI text templates}. We conducted additional ablation studies to investigate various components of FreeA, and the results have been tabulated in Table \ref{tab-modules}. 
It shows that both HOI text templates $T_1$ and $T_2$ play a vital role in HOI detection, at a 0.93 mAP increase as compared to FreeA with $T_1$ (Rows 1 and 2). This is mainly because we observe that $T_1$ text template (``a photo of a person verb-ing an object) does not capture significant differences in similarity between different actions on the same object. Therefore, we introduce the $T_2$ text template, which emphasizes the action (``a photo of a person verb-ing").

\begin{figure*}[ht]
\vspace{-0.3cm}
\centering 
\includegraphics[width=\linewidth]{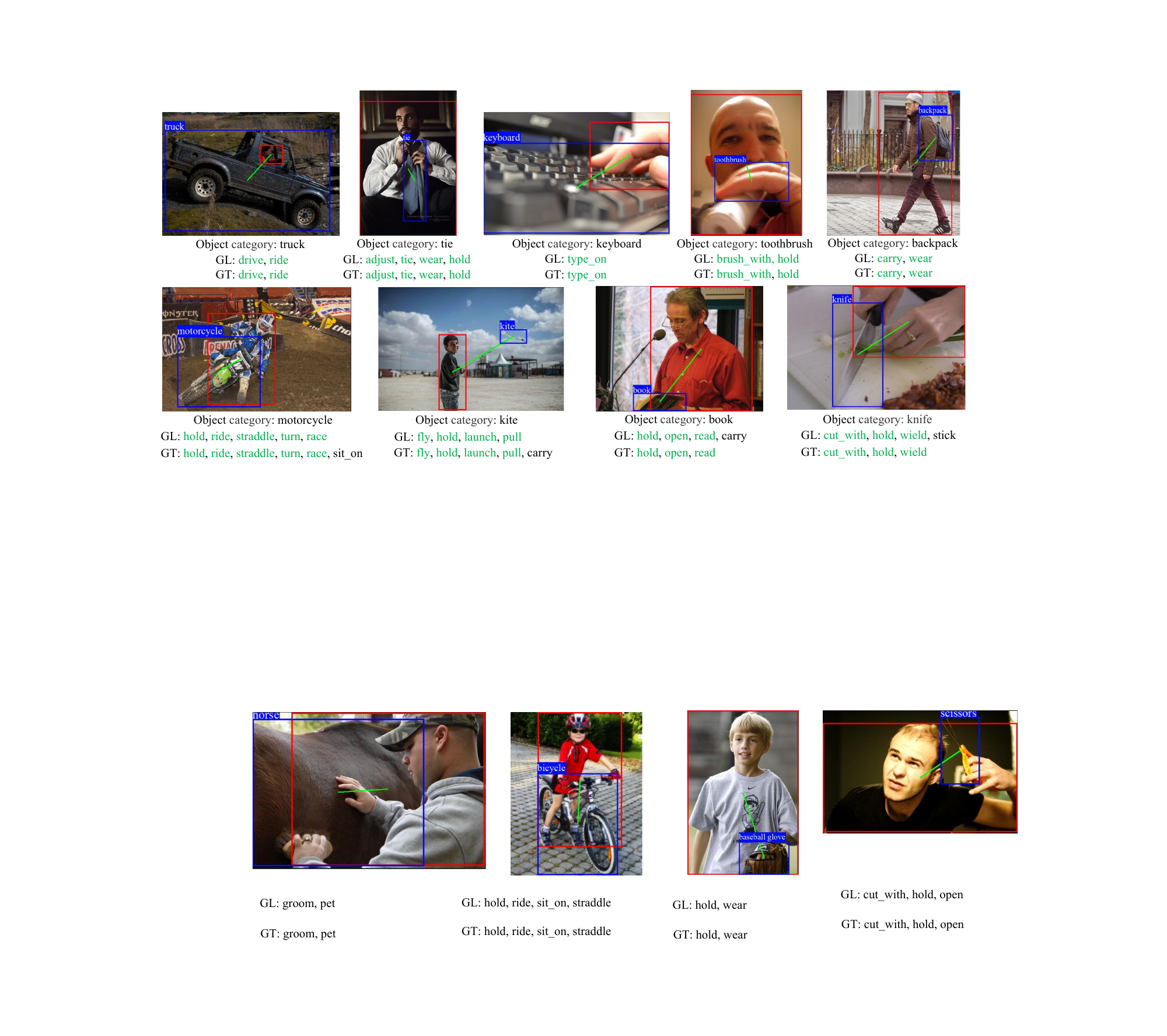}
\caption{Comparison of HOI labels. GL and GT represents generated labels and ground truth, respectively. The red and blue rectangles are bounding boxes for the human and object, and the green lines represent the connection between their centers. Green text indicates correct interactions.}
\label{fig:label_results}
\vspace{0.2cm}
\end{figure*}

\begin{table*}[h]
\centering
\caption{Ablation study using different modules on HICO-Det datasets using Swin-L backbone and Yolov8 detector under weakly+ settings. RB denotes retaining the background. DB indicates deleting the background.}
\label{tab-modules}
\setlength{\tabcolsep}{1.0mm}{
\begin{tabular}{ccccccccccc}
\specialrule{1.5pt}{0pt}{0pt}
Row & $T_1$   & $T_2$   & $top$-1 & Adaption & \begin{tabular}[c]{@{}l@{}}Dynamic\\ threshold\end{tabular} & \begin{tabular}[c]{@{}c@{}}Segmentation\\ (RB)\end{tabular} & \begin{tabular}[c]{@{}c@{}}Segmentation\\ (DB)\end{tabular} & ICM  & PKM   & \begin{tabular}[c]{@{}l@{}}\textbf{mAP}$\uparrow$\\ (Full)\end{tabular} \\ \hline
1   & $\checkmark$ &         & $\checkmark$ &          &                                                              & $\checkmark$                                                                   &                                                                          &   &       & 25.83                                                \\
2   & $\checkmark$ & $\checkmark$ & $\checkmark$ &          &                                                              & $\checkmark$                                                                   &                                                                          &       &  & 26.76                                               \\
3   & $\checkmark$ & $\checkmark$ &         & $\checkmark$  &                                                              & $\checkmark$                                                                   &                                                                          &     &    & 27.85                                                \\
4   & $\checkmark$ & $\checkmark$ &         & $\checkmark$  & $\checkmark$                                                      & $\checkmark$                                                                   &                                                                          &    &     & 29.03                                                \\
5   & $\checkmark$ & $\checkmark$ &         & $\checkmark$  & $\checkmark$                                                      &                                                                           & $\checkmark$                                                                  &     &    & 29.94                                                \\
6   & $\checkmark$ & $\checkmark$ &         & $\checkmark$  & $\checkmark$                                                      &                                                                           & $\checkmark$                                                                  & $\checkmark$ & & 31.48                                                \\ 
\rowcolor[gray]{0.9}7   & $\checkmark$ & $\checkmark$ &         & $\checkmark$  & $\checkmark$                                                      &                                                                           & $\checkmark$                                                                  & $\checkmark$ & $\checkmark$ &\textbf{33.01}                                                \\ 
\specialrule{1.5pt}{0pt}{0pt}
\end{tabular}}

\end{table*}

\textbf{Action selection approaches}. Two action selection approaches, namely, ``$top$-1" and ``adaption", are designed to determine which action should be selected when human-object interactions are not present in an image (Row 2 and Row 3 in Table \ref{tab-modules}).
The ``$top$-1" refers to selecting the most salient action, whereas ``adaption" (Eq. \ref{eq:adapt}) retains actions within a specified threshold range. The results show that the ``adaption'' approach outperforms the ``$top$-1" approach at +1.09 mAP. 
The $top$-1 only selects the action with the highest similarity, however, HOI typically involves interactions with multiple actions, and the ``adaptation" method can satisfy this to provide multiple choices.

\textbf{Dynamic threshold}. We further test the effect of dynamic threshold $\theta$ (Eq. \ref{eq:dynamic-weight}) on the FreeA (Row 3 and Row 4 in Table \ref{tab-modules}). It is observed that using a dynamic threshold instead of a fixed threshold results in a 1.18 mAP improvement. This indicates that the variability in the subtraction of the average similarity of all actions from the highest similarity action obtained through text-image matching model is significant, and a fixed threshold cannot overcome this problem. 

\textbf{Background retention or deletion}. The image background can be a double-edged sword. Sometimes it works in your favor for simple image background, sometimes it works against you when one includes complex background details leading to different interferences. We conducted experiments to verify the effect of image background. The results indicate that retaining the background leads to a performance decrease (Row 4 and Row 5 in Table \ref{tab-modules}).

\textbf{ICM and FCM}. Regarding the proposed ICM component, the experimental results demonstrate an improvement of 1.54 mAP when ICM is utilized compared to when it is not (Row 5 and Row 6 in Table \ref{tab-modules}). The improvement is foreseeable because the ICM module emphasizes the correlation between actions. Furthermore, comparing rows 6 and 7 in Table~\ref{tab-modules}, we observe an additional gain of 1.53 mAP after incorporating the PKM module, 
demonstrating its effectiveness in enhancing performance by leveraging prior knowledge of object-action associations. 

\section{Visualization}
We visualize some results of generated labels compared with ground truth HOI labels, as shown in Figure \ref{fig:label_results}. It’s clear that the generated labels have a high overlap with the ground truth labels. 


\section{Conclusion}
We propose a novel weakly\text{-}\text{-} supervised HOI detection method, termed FreeA. Weakly\text{-}\text{-} supervised FreeA means the training labels are not manually annotated from the raw datasets, but automatically generated from the text-image matching model with the combination of candidate image construction, human-object potential interaction mining, and human-object interaction inference modules. Compared with those weakly, weakly+, and fully supervised HOI methods, extensive experiments have demonstrated the effectiveness and advantages of the proposed FreeA. Our contributions to the field include presenting a new problem of weakly supervised HOI detection and showing the utilization of the text-image model for generating HOI labels. 

{
\small
\bibliographystyle{plainnat}
\bibliography{ref}
}

\clearpage

\end{document}